%% file: emnlp2021.tex
\pdfoutput=1

\documentclass[11pt]{article}

\usepackage[]{emnlp2021}

\usepackage{times}
\usepackage{latexsym}
\usepackage{amsmath}
\usepackage{graphicx}  
\usepackage[T1]{fontenc}

\usepackage{graphicx,subfigure}

\usepackage{enumitem}
\setlist[itemize,enumerate]{leftmargin=*}

\usepackage[T2A,T1]{fontenc}
\usepackage[utf8]{inputenc}

\usepackage{dsfont}
\usepackage{microtype}
\usepackage{amssymb}
\usepackage{multirow}
\usepackage{color, colortbl}
\usepackage{balance}
\usepackage{booktabs}
\usepackage[normalem]{ulem}

\newcommand{\edit}[1]{\textcolor{black}{#1}}

\definecolor{royalblue}{rgb}{0.25, 0.41, 0.88}
\definecolor{internationalorange}{rgb}{1.0, 0.31, 0.0}
\definecolor{green(ncs)}{rgb}{0.0, 0.62, 0.42}
\definecolor{Gray}{gray}{0.9}
\definecolor{LightCyan}{rgb}{0.88,1,1}
\definecolor{palecornflowerblue}{rgb}{0.67, 0.8, 0.94}

\title{Uncertainty-Aware Machine Translation Evaluation}

\author{
Taisiya Glushkova$^{1, 4}$ 
Chrysoula Zerva$^{1, 4}$  
Ricardo Rei$^{2, 3, 4}$  
André F. T. Martins$^{1, 2, 4}$ 
\\
$^1$Instituto de Telecomunicações \quad $^2$Unbabel \quad $^3$INESC-ID \\
$^4$Instituto Superior Técnico \& LUMLIS (Lisbon ELLIS Unit) \\
{\small \texttt{\{taisiya.glushkova, chrysoula.zerva, andre.t.martins\}@tecnico.ulisboa.pt}}\\
{\small \texttt{ricardo.rei@unbabel.com}
}}

\begin{document}
\maketitle
\begin{abstract}



Several neural-based metrics have been recently proposed to evaluate machine translation quality. However, all of them resort to point estimates, which provide limited information at segment level. This is made worse as they are trained on noisy, biased and scarce human judgements, often resulting in unreliable quality predictions.
In this paper, we introduce \textit{uncertainty-aware} MT evaluation and analyze the trustworthiness of the predicted quality.
We combine the \textsc{Comet} framework with two uncertainty estimation methods, Monte Carlo dropout and deep ensembles, to obtain quality scores along with confidence intervals. 
We compare the performance of our uncertainty-aware MT evaluation methods across multiple language pairs from the QT21 dataset and the WMT20 metrics task, augmented with MQM annotations. We experiment with varying numbers of references and further discuss the usefulness of uncertainty-aware quality estimation (without references) to flag possibly critical translation mistakes.




\end{abstract}

\section{Introduction}

\input{table_examples.tex}

Evaluation of machine translation (MT) quality is a key problem with several use cases: it is needed to compare and select MT systems, to decide on the fly whether a translation is ready for publication or needs to be post-edited by a human, and more generally to track progress in the field \citep{specia2018quality,mathur-etal-2020-results}. 
Even when reference translations are available, the increasing quality of neural MT systems has made traditional lexical-based metrics such as \textsc{Bleu} \cite{papineni-etal-2002-bleu} or \textsc{chrF} \cite{popovic-2015-chrf} insufficient to distinguish the best systems. This fostered a line of work on neural-based metrics, with recent proposals such as \textsc{Bleurt} \cite{sellam-etal-2020-bleurt}, \textsc{Comet} \cite{rei-etal-2020-comet} and \textsc{Prism} \cite{thompson-post-2020-automatic}. Metrics for quality estimation (QE; when references are not available) have also been developed as part of \textsc{OpenKiwi} \cite{kepler-etal-2019-openkiwi} and \textsc{TransQuest} \citep{ranasinghe-etal-2020-transquest}. 

While the metrics above have enjoyed some success in system-level evaluation -- where the goal is to compare different systems -- 
their segment-level quality scores are often unreliable for practical use. 
They all share the limitation that their output is a single \emph{point estimate} -- they do not provide any uncertainty information, such as confidence intervals, with their quality predictions. This is an important limitation: often, complex or out-of-domain sentences receive quality estimates that are far from their true quality (as illustrated in Table~\ref{tab:examples}). This may lead to translations with critical mistakes being undetected, and hinders worst-case performance analysis of MT systems. 

In this paper, we propose a simple and effective method to obtain \textbf{uncertainty-aware} quality/metric estimation systems, by representing quality as a \emph{distribution}, rather than a single value. 
To this end, we make use of and compare two well-studied techniques for uncertainty estimation: Monte Carlo (MC) dropout \cite{gal2016dropout} and deep ensembles \cite{lakshminarayanan2017simple}. In both cases, our method is agnostic to the particular metric estimation system, as long as it can be ensembled or perturbed. In our experiments we use \textsc{Comet} \cite{rei-etal-2020-comet}, and we call our uncertainty-aware version \textsc{UA-Comet}.%
\footnote{Link to our code can be found at \url{https://github.com/deep-spin/UA_COMET}. A newer version of COMET, with incorporated uncertainty options is available at \url{https://github.com/Unbabel/COMET}.} %

Our method allows using the same system with a varying number of references. 
We show that confidence intervals tend to shrink as more references are added, which matches the intuition that MT evaluation systems should become more confident as they have access to more information. 

We evaluate our approach using data from the WMT20 metrics task \citep{mathur-etal-2020-results}, including its recent extension with Google MQM annotations \citep{freitag2021experts}, and the QT21 dataset \cite{specia2017translation}. The results show that our uncertainty-aware systems exhibit better calibration with respect to human direct assessments (DA; \citealt{graham-etal-2013-continuous}), multi-dimensional quality metric scores (MQM; \citealt{lommel2014multidimensional}), and human translation error rates (HTER; \citealt{snover2006study})
than a simple baseline, while their average quality scores achieve similar or better correlation than the vanilla \textsc{Comet} system. Finally, we illustrate a potential quality estimation use case enabled by our approach: automatically detecting low-quality translations with a risk-based criterion.



\section{Related Work}\label{sec:related}


\paragraph{Automatic MT evaluation}

Reference-based approaches for MT evaluation include traditional metrics such as \textsc{Bleu}~\cite{papineni-etal-2002-bleu} and \textsc{Meteor}~\cite{denkowski-lavie-2014-meteor}, as well as recently proposed  \textsc{Bleurt}~\cite{sellam-etal-2020-bleurt}, \textsc{BERTScore}~\cite{zhang2019bertscore}, \textsc{Prism}~\cite{thompson-post-2020-automatic} and \textsc{Comet}~\cite{rei-etal-2020-comet}. Approaches that do not make use of human references are generally referred to as QE systems \cite{specia2018quality,kepler-etal-2019-openkiwi,ranasinghe-etal-2020-transquest}. Our proposed approach augments reference-based approaches and enables a single system that can be used with multiple references, with the added advantage of providing uncertainty information. 
To the best of our knowledge, predictive uncertainty in QE has been approached only with Gaussian processes 
\cite{beck-etal-2016-exploring}, which are not competitive or easy to integrate with current neural architectures.



\paragraph{Confidence estimation in MT}

A related line of work is confidence estimation of sentence-level MT outputs \cite{blatz-etal-2004-confidence, quirk-2004-training, wang-etal-2019-improving-back}. The work that relates the most with ours is the one by \citet{fomicheva-etal-2020-unsupervised}, who propose an unsupervised glass-box approach to QE, extracting uncertainty-related features from the MT system via MC dropout. 
They show that the more confident the decoder (as measured by the lower variance of its output), the higher the quality of the MT output. 
Our work builds upon this perspective to propose uncertainty estimation of the QE systems themselves, rather than uncertainty of MT. 

\paragraph{Performance prediction in NLP} 
A related problem is that of predicting the performance of an NLP system without having to train it \citep{xia-etal-2020-predicting}. Recent approaches perform such predictions by adding confidence intervals \citep{ye-etal-2021-towards} and measuring calibration error. 
We take inspiration from these works to improve the calibration of our methods \cite{guo2017calibration,desai-durrett-2020-calibration} and to evaluate how good our uncertainty estimates are with a suite of performance indicators. 

\paragraph{Uncertainty estimation}
Overall the concepts and methods of uncertainty quantification \cite{hullermeier2019aleatoric} have been widely explored and compared for many different tasks, 
including MT \cite{ott2018analyzing}. Uncertainty estimation in neural networks has traditionally been approached with Bayesian methods, replacing point estimates of weights with probability distributions \cite{mackay1992bayesian, graves2011practical, welling2011bayesian, tran2019bayesian}. However, Bayesian neural networks are costly, and in order to avoid high training costs, various approximations come in handy. Model ensembling \cite{dietterich2000ensemble, garmash-monz-2016-ensemble, mcclure2016robustly, lakshminarayanan2017simple, pearce2018uncertainty, jain2020maximizing} is a commonly used approach, which employs an ensemble of neural networks to obtain multiple point predictions and then uses their empirical variance as an approximate measure of uncertainty. Its main disadvantage is the need to train multiple models. 
An alternative is MC dropout \citep{gal2016dropout}, which builds upon dropout regularization \citep{srivastava2014dropout} but uses it at test time,  by performing several stochastic forward passes through the network and computing mean and variance of the resulting outputs as a proxy for the model's uncertainty. 
Our work applies and compares the last two techniques to MT evaluation. Note that more elaborate approaches have been proposed to address uncertainty quantification on classification tasks, including calibration approaches \cite{guo2017calibration,pmlr-v80-kuleshov18a}, the use of Dirichlet distributions \cite{sensoy2018evidential,malinin2018predictive,charpentier2020posterior} and entropy measures \cite{smith2018understanding}. However, uncertainty in MT evaluation is a  regression task which is so far largely overlooked in terms of predictive uncertainty. Our paper can be seen as a first step towards uncertainty-aware MT evaluation models.

\section{Uncertainty-Aware MT Evaluation}\label{sec:uncertainty}




\subsection{Problem definition}

Typical MT evaluation systems take as input a tuple $\langle s, t, \mathcal{R}\rangle$, where $s$ is source text, $t$ is machine translated text, and $\mathcal{R} = \{r_1, \ldots, r_{|\mathcal{R}|}\}$ is a (possibly empty) set of reference translations. 
Their goal is to predict an automatic score $\hat{q} \in \mathbb{R}$ which assesses the quality of the translation. %
Supervised systems such as \textsc{Comet} or \textsc{Bleurt} are 
trained to approximate  ground truth scores $q^*$ obtained from human annotations, such as 
DA, MQM and HTER. In this paper, we assume that $q^*$ is a continuous real-valued score, but the main ideas extend to the case where $q^*$ are discrete classes or quality bins. 

\subsection{Sources of uncertainty}
\label{sec:unc_sources}
There are several challenges with learning MT evaluation systems:
\begin{enumerate}
\item \textbf{Noisy scores.} The human-generated scores $q^*$ are not always reliable and often suffer from high variability, exhibiting low inter-annotator agreement. This problem can be mitigated by averaging over  a sufficient number of references, but this brings considerable annotation costs \cite{freitag2021experts, mathur-etal-2020-results}. 
\item \textbf{Noisy or insufficient references.} The references $\mathcal{R}$ do not always have good quality, and their sparsity (small $|\mathcal{R}|$) is often insufficient to represent the space of possible correct translations well~\cite{freitag-etal-2020-bleu}.%
\footnote{From the perspective of the MT system, the existence of multiple valid translations for a single source sentence can be seen as \textit{inherent uncertainty} of the task \citep{ott2018analyzing}.} %
An extreme case is when there are no references ($\mathcal{R}=\varnothing$), a problem known as ``QE as a metric.''
\item \textbf{Complex translations.} Correct translations are often non-literal, and it may be hard for an automatic system to grasp the semantic relation between the translated sentence and the references, as they may be confused with hallucinations. 
\item \textbf{Out-of-domain text.} The text where the MT evaluation system is run may belong to a different domain from the one it was trained on.
\end{enumerate} 
The first two points can be seen as \textit{aleatoric} uncertainty (noise in the input or output data), whereas the last two are instances of \textit{epistemic} uncertainty, reflecting the limited knowledge of the model \cite{hora1996aleatory, der2009aleatory}. Unfortunately, these uncertainties add up. 
To cope with the different sources of uncertainty, we treat the quality score $Q$ as a random variable and predict a \textbf{distribution} $\hat{p}_Q(q)$, as opposed to a point estimate $\hat{q}$. 
This way, we obtain an \textbf{uncertainty-aware} system, which can return a peaked distribution when it is confident about its quality estimate, or a flatter distribution in cases where it is more uncertain. 
This allows, among other things, managing the risk of treating a translation as good quality when it is not (see \S\ref{sec:multi-ref-exp}).  
When estimating quality on the fly without references, knowing the system's confidence in the quality of the produced translations might help obtain informative worst-case indicators on whether a human post-edit is required, e.g. by evaluating the cumulative distribution function $\hat{F}_Q(\chi) = \int_{-\infty}^\chi \hat{p}_Q(q) dq$ which quantifies the \textbf{translation risk}, i.e., the probability of a translation being below a quality threshold $\chi$. 
Moreover, having access to such distributions of quality estimates 
can be beneficial when deciding if a system outperforms another with some level of confidence.  

\subsection{Uncertainty and confidence intervals}





To obtain $\hat{p}_Q(q)$, our approach builds upon a vanilla MT evaluation system $h$ (such as $\textsc{Comet}$) that produces point estimates $\hat{q} = h(\langle s, t, \mathcal{R} \rangle)$, and \textit{augments it} to produce uncertainty estimates. Our approach is  completely agnostic about the 
system $h$, as long as it can be ensembled or perturbed. 

The first step is to use $h$ to produce a set $\mathcal{Q} = \{\hat{q}_1, \ldots, \hat{q}_N\}$ of quality scores for a given input $\langle s, t, \mathcal{R} \rangle$, which will be interpreted as a sample from $\hat{p}_Q(q)$. For this, 
we experiment with two methods: \textbf{MC dropout} \citep{gal2016dropout}, which obtains $\mathcal{Q}$ by running $N$ stochastic forward-passes on $h$ with units dropped out with a given probability; 
and \textbf{deep ensembles} \citep{lakshminarayanan2017simple}, in which $N$ separate models are trained with different random initializations and then run in parallel to obtain $\mathcal{Q}$. 
While both methods have shown to be effective in several tasks \citep{fomicheva-etal-2020-unsupervised, malinin2020uncertainty}, 
MC dropout is usually more convenient (because only one model is required), but generally requires many more samples for good performance (larger $N$) compared to deep ensembles. 

The second step is to use the resulting set $\mathcal{Q}$ to represent model's uncertainty. 
One way of representing uncertainty is through \textbf{confidence intervals}, that is, given a desired confidence level $\gamma \in [0,1]$ (e.g. $\gamma = 0.95$), specifying the smallest possible quality interval $I(\gamma) = [q_{\min}(\gamma), q_{\max}(\gamma)]$  such that $P(q \in I(\gamma)) = \int_{q_{\min}}^{q_{\max}}\hat{p}_Q(q) dq \ge \gamma$. 
There are two possible strategies to obtain such intervals: a \textit{parametric} approach, which parametrizes the distribution $\hat{p}_Q(q)$, produces estimates of its parameters by fitting the distribution on $\mathcal{Q}$, and uses them to compute confidence intervals at arbitrary levels $\gamma$; 
and a \textit{non-parametric} approach, which bypasses the estimation of $\hat{p}_Q(q)$ and focuses on estimating its quantiles for the desired values of $\gamma$ directly from $\mathcal{Q}$. 
In this paper, we opted for a simple parametric Gaussian approach, which worked well in practice and seemed to fit our data well (see Figure~\ref{fig:gaussian-data} in  App.~\ref{sec:datasets}).  
However, we did experiment with a non-parametric bootstrapping technique using the percentile method \citep{efron1992bootstrap,johnson2001introduction,ye-etal-2021-towards}, which we report in App.~\ref{sec:non-parametric}.

In our approach, we treat $\mathcal{Q}$ as a sample drawn from a Gaussian distribution, $\hat{p}_Q(q) = \mathcal{N}(q; \hat{\mu}, \hat{\sigma}^2)$, and estimate the parameters  $\hat{\mu}$ and $\hat{\sigma}^2$ as the sample mean and variance, respectively. 
Once $\hat{p}_Q(q)$ is fit to $\mathcal{Q}$, the confidence intervals $I(\gamma) = [q_{\min}(\gamma), q_{\max}(\gamma)]$ can be estimated at the desired level of confidence $\gamma$, using the probit (quantile) function $\mathrm{probit}(p) = \sqrt{2}\mathrm{erf}^{-1}(2p-1)$ (where $\mathrm{erf}$ is the error function):
\begin{align}
    q_{\min}(\gamma) &= \hat{\mu} - \hat{\sigma} \mathrm{probit}((1+\gamma)/2)\nonumber\\
    q_{\max}(\gamma) &= \hat{\mu} + \hat{\sigma} \mathrm{probit}((1+\gamma)/2).
\end{align}

\subsection{MT evaluation with multi-references}
\label{sec:refs}
As our framework can model uncertainty, it is interesting to consider the case where the number of available references $\mathcal{R}$ may vary. Intuitively, we expect the uncertainty to decrease when the model observes more references. Specifically, relying on a single reference might prove problematic, since even human generated references can be noisy and prone to errors. Additionally, for source sentences with multiple and diverse valid translations, relying on a single reference might result in potential underestimation of the quality of valid MT hypotheses. 
For the above reasons, additional references, even if they are paraphrased versions of the originals \cite{freitag-etal-2020-bleu}, can help obtain better evaluations of the MT systems' outputs. 

As a result, relying on human-generated references can be a constraint in terms of learning and predicting accurate quality estimates for adequately diverse data \cite{sun-etal-2020-estimating}. We thus want to assess the impact of additional references (both independently generated and paraphrased) on the estimated confidence intervals. 

Even though our approach works with any underlying MT evaluation system $h$ which produces point estimates, most existing systems cannot seamlessly handle a varying number of references or no references without architecture modifications. For example, \textsc{Comet} originally receives exactly one reference as input to predict the quality of a $\langle s, t \rangle$ pair. We take the following approach to handle a varying number of references ($|\mathcal{R}| > 1$):  we obtain a set of $N$ quality predictions for each available reference, $r \in \mathcal{R}$, for a given $\langle s, t \rangle$ pair, resulting in a set of $N \times |\mathcal{R}|$ quality predictions. We then compute the pointwise average across the $|\mathcal{R}|$ dimension, leading to $N$ quality scores $\mathcal{Q} = \{\hat{q}_1, \ldots, \hat{q}_N\}$ that aggregate information from all the $|\mathcal{R}|$ references. We can then apply the same approach as described earlier. Intuitively, the averaging operation should reduce variance in the quality scores, which would result in narrower confidence intervals as $|\mathcal{R}|$ increases.  
We validate this hypothesis in our experiments in \S\ref{sec:multi-ref-exp}.

\subsection{Post-calibration}
\label{sec:postcal}

\begin{figure}[t]
\centering
\includegraphics[width=\columnwidth]{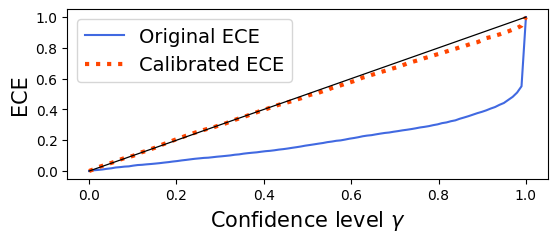}
\caption{Calibration for \textsc{En-De} language pair. Continuous (\textcolor{royalblue}{\textbf{blue}}) line is ECE pre-calibration and dotted (\textcolor{internationalorange}{\textbf{orange}}) line corresponds to ECE post-calibration.}
\label{fig:calibration-plot}
\end{figure}

In our initial experiments, we observed that
the magnitude of the predicted variance $\hat{\sigma}^2$ depends significantly on several hyperparameters, such as the choice of dropout value, number of samples, and language pair. 
In classification tasks, a similar phenomenon has been reported by \citet{malinin2020uncertainty}, who recommended combining these methods with temperature calibration \cite{platt1999probabilistic} to adjust uncertainties and obtain more reliable confidence intervals.  
For regression tasks -- our case of interest -- \citet{kuleshov2018accurate} also point out the importance of post-calibration. Since temperature scaling is only applicable in classification, they propose an isotonic regression technique instead \cite{niculescu2005predicting}. 
We found that we can obtain highly calibrated uncertainty estimates in a much simpler way, by learning an affine transformation $\sigma^2 \mapsto \sigma^2_{\mathrm{calib}} = \alpha \sigma^2 + \beta$, where $\alpha$ and $\beta$ are scalars, tuned to minimize the calibration error (see Eq. \ref{eq:ece}--\ref{eq:ece_acc}) on a 
validation set. 
We use the tuned  $\sigma_{\mathrm{calib}}$ in our experiments (\S\ref{sec:experiments}), and show the improvement on ECE for different confidence levels with $\sigma_{\mathrm{calib}}$ in Figure~\ref{fig:calibration-plot}. 

\section{Evaluating Uncertainty}\label{sec:evaluation}

Having described our framework, we now turn to the problem of verifying the effectiveness and informativeness of the proposed uncertainty quantification method. 
Two crucial aspects to take into account when evaluating uncertainty-aware systems are: (i) the system should not harm the predictive accuracy compared to a system without uncertainty and (ii) the uncertainty estimate should reflect the failure probability of the system well, meaning that the system ``knows when it does not know.'' 
In what follows, we assume a test or validation set $\mathcal{D} = \{\langle s_j, t_j, \mathcal{R}_j, q_j^* \rangle\}_{j=1}^{|\mathcal{D}|}$, consisting of examples together with their ground truth quality scores.

\paragraph{Calibration Error} 
One way of understanding if models can be trusted is analyzing whether they are \textit{calibrated} \cite{raftery2005using, jiang2012calibrating, kendall2017uncertainties}, that is, if the confidence estimates of its predictions are aligned with the empirical likelihoods \cite{guo2017calibration}. In classification tasks, this is assessed by the \textit{expected calibration error} (ECE; \citealt{naeini2015obtaining}), which has been generalized to regression by \citet{kuleshov2018accurate}. 
It is defined as:
\begin{align}
\label{eq:ece}
    \mathrm{ECE} &= \frac{1}{M}\sum_{b=1}^{M} |\mathrm{acc}(\gamma_{b}) - \gamma_{b}|,
\end{align}
where each $b$ is a bin representing a confidence level $\gamma_b$, and $\mathrm{acc}(\gamma_{b})$ is the fraction of times the ground truth $q^*$ falls inside the confidence interval $I(\gamma_b)$:
\begin{equation}
\label{eq:ece_acc}
    \mathrm{acc}(\gamma_{b}) = \frac{1}{|\mathcal{D}|} \sum_{\langle s, t, \mathcal{R}, q^* \rangle \in \mathcal{D}} \mathds{1}(q^* \in I(\gamma_b)).
\end{equation}
We use this metric with $M=100$. 

\paragraph{Negative log-likelihood} 

To evaluate parametric methods that represent the full distribution $\hat{p}_Q(q)$, we can use a single metric that captures both accuracy and uncertainty, the average negative log-likelihood of the ground truth quality scores according to the model:
\begin{equation}
    \mathrm{NLL} = -\frac{1}{|\mathcal{D}|}\sum_{\langle s, t, \mathcal{R}, q^* \rangle \in \mathcal{D}} \log \hat{p}(q^* \mid \langle s, t, \mathcal{R} \rangle).
\end{equation}
This metric penalizes predictions that are accurate but have high uncertainty (since they will become flat distributions with low probability everywhere), and even more severely incorrect predictions with high confidence (as they will be peaked in the wrong location), but is more forgiving to predictions that are inaccurate but have high uncertainty.

\paragraph{Sharpness} 
The metrics above do not sufficiently account for how ``tight'' the uncertainty interval is around the predicted value, and thus might generally favour predictors that produce wide and uninformative confidence intervals. 
To guarantee useful uncertainty estimation, confidence intervals should not only be calibrated, but also sharp. 
We measure sharpness using the predicted variance $\hat{\sigma}^2$, as defined in \citet{kuleshov2018accurate}:

\begin{equation}
    \mathrm{sha}(\hat{p}_Q) = \frac{1}{|\mathcal{D}|} \sum_{\langle s, t, \mathcal{R} \rangle \in \mathcal{D}}  \hat{\sigma}^2.
\end{equation}

\paragraph{Pearson correlations} 
\label{sec:pearson_correlations}
As shown by \citet{ashukha2019pitfalls}, NLL and ECE alone might not be enough to evaluate uncertainty-aware systems. 
Therefore, we complement the indicators above with two Pearson correlations involving the system's predictions and the ground truth quality scores coming from human judgements. 
The first, which we call the \textbf{predictive Pearson score} (PPS), is useful to assess the predictive accuracy of the system, regardless of the uncertainty estimate -- it is the Pearson correlation $r(q^*, \hat{\mu})$ between the ground truth quality scores $q^*$ and the average system predictions $\hat{\mu}$ in the dataset $\mathcal{D}$ (for the  
baseline point estimate system, we use $\hat{q}$ instead of $\hat{\mu}$). We expect this score to be similar to the baseline or slightly better due to the ensemble effect. 
The second is the \textbf{uncertainty Pearson score} (UPS) $r(|q^*-\hat{\mu}|, \hat{\sigma})$, which measures the alignment between the prediction errors $|q^* - \hat{\mu}|$ and the uncertainty estimates $\hat{\sigma}$. 
Note that achieving a high UPS is much more challenging -- a model with a very high score would know how to correct its own predictions to obtain perfect accuracy. We confirm this claim later in our experiments.

\section{Experiments}
\label{sec:experiments}
\subsection{Datasets}
\label{sec:data}

We apply our method to predict three types of human judgement scores at segment-level: DA, MQM and HTER.
We use the WMT20 metrics shared task dataset \cite{mathur-etal-2020-results} for the DA judgements, and the Google MQM annotations for English-German (\textsc{En-De}) and Chinese-English (\textsc{Zh-En}) on the same corpus~\cite{freitag2021experts}. For language pairs where both human- and system- generated translations are provided, we remove the human translations before evaluating (Human-A, Human-B, Human-P in WMT20).
For the HTER experiments, we use the QT21 dataset   \cite{specia2017translation}. Dataset statistics are presented in App.~\ref{sec:datasets}.

\subsection{Experimental setup}
\label{sec:exp_setup}

For the experiments presented below, we use \textsc{Comet} as the underlying MT quality evaluation system \citep{rei-etal-2020-comet}.%
\footnote{More precisely we used the \textit{wmt-large-da-estimator-1719} and the \textit{wmt-large-hter-estimator} available at: \url{https://unbabel.github.io/COMET/html/models.html}.} %
For evaluation, 
we perform $k$-fold cross-validation:  
we split the test partition 
into $k = 5$ folds, so that each fold contains translations of every MT system and has approximately the same number of documents. 
The $k$-fold splits are generated in such a way that there are unique source-reference pairs in each fold, and the documents are disjunct across the folds. Since documents vary in their length, the number of segments per fold can differ. We use 4 folds for validation and the remaining one for testing. 
As we experiment with human annotations of different scales, $\hat{q}$ and $q^*$ are standardized on the validation set and the model is post-calibrated as described in \S\ref{sec:postcal}. 


\paragraph{MC dropout (MCD)} We apply a dropout probability of 0.1 and run $N=100$ runs of MC dropout. Dropout was applied at encoder, pooling and feed-forward layers as we found it produces more useful $\hat{\sigma}$ values, corroborating the findings of \citet{verdoja2020notes} and \citet{kendall2017bayesian}. 
More details on tuning the hyperparameters can be found in App.~\ref{sec:hyperparams}.


\paragraph{Deep Ensembles (DE)} We train ensembles with $N=5$ models and random initialization. For training, we follow the procedure described by \citet{rei-etal-2020-unbabels}, training each model for $2$ epochs. 

\paragraph{Baseline}
As a simple baseline, we take the original point estimates $\hat{q}$ provided by the underlying \textsc{Comet} system and map them to a Gaussian distribution $\mathcal{N}(q; \hat{\mu}, \hat{\sigma}^2)$ with $\hat{\mu} := \hat{q}$ and a fixed variance $\hat{\sigma}^2 := \sigma_{\mathrm{fixed}}^2$ (i.e., the same variance is assigned to all the examples). We compute $\sigma_{\mathrm{fixed}}^2$ on the validation set so that it minimizes the average NLL value, which has the following closed form expression (see App.~\ref{sec:optimal_fixed_variance} for a proof):
\begin{equation}
\label{eq:baseline}
    \sigma_{\mathrm{fixed}}^2 = \frac{1}{|\mathcal{D}|} \sum_{{\langle s, t, \mathcal{R}, q^* \rangle \in \mathcal{D}}} (q^* - \hat{\mu})^2 .
\end{equation}
This baseline was found surprisingly strong on several performance indicators (Tables~\ref{tab:da_seg},~\ref{tab:mqm_seg},~\ref{tab:hter_seg}).


\subsection{Segment-level analysis}
Table~\ref{tab:da_seg} presents results for the  performance indicators described in \S\ref{sec:evaluation} for 9 language pairs in the WMT20 dataset, encompassing a mix of high-resource and low-resource languages.  
We observe that both uncertainty-aware methods (MCD and DE) show consistent improvement over the baseline in all metrics and language pairs, with the exception of NLL in two language pairs (\textsc{Zh-En} and \textsc{En-Iu}).
We also see that, overall, deep ensembles provide more accurate predictions and narrower confidence intervals compared to MC dropout, but without a significant improvement for the other performance indicators across pairs. Considering the computational cost of training and tuning multiple models for the deep ensemble, MC dropout seems preferable for the presented MT evaluation setup. 

While these results are encouraging, we stress that experiments on higher quality data at a larger scale are necessary to fully validate and compare uncertainty-aware methods, 
as the numbers in Table ~\ref{tab:da_seg} are influenced by the inconsistencies in DA annotations, which are known to be particularly noisy \cite{toral-2020-reassessing,freitag2021experts}. 
To mitigate this, we further compare performance on the recently released Google MQM annotations for \textsc{En-DE} and \textsc{Zh-En}, shown in Table~\ref{tab:mqm_seg}. 
As expected from the higher quality of these annotations, and even though the underlying \textsc{Comet} system was still trained on DAs and evaluated on the MQM assessments, 
we get higher uncertainty correlations, with the MC dropout approach benefiting the most. We also notice a significant improvement across all indicators for the \textsc{Zh-En} dataset, which was poorly correlated with the predictions on the DA dataset.
We use the MQM annotations to provide a more in-depth analysis on specific use cases on translation evaluation in \S\ref{sec:multi-ref-exp} -\ref{sec:ir}. 

Finally, Table~\ref{tab:hter_seg} shows the results on HTER prediction on the QT21 dataset.%
\footnote{This dataset contains post-edits of the MT output, for which the HTER score is computed, and independent human references, which we use to predict HTER following the same experimental procedure as \citet{rei-etal-2020-comet}.} %
For this metric and dataset, the Pearson correlations are generally higher than in previous experiments (with the exception of UPS for \textsc{En-Cs}) and the sharpness scores indicate that the predicted confidence intervals are considerably narrower, showing that for these experiments the models are generally more accurate and more confident than when predicting DA and MQM. This might be explained by the fact that HTER, which quantifies the amount of post-editing required to fix a translation, is a less subjective metric than a quality assessment, and therefore the aleatoric uncertainty caused by noisy scores may be smaller.

\input{table_WMT20}

\input{table_MQM}


\input{table_QT21}

\subsection{Impact of reference quantity}
\label{sec:multi-ref-exp}
We next experiment with the WMT20 \textsc{En-De} to get some insights on the impact of using multiple references as described in \S\ref{sec:refs}. 
This dataset contains 3 human references (Human A, B, and P) for each source sentence generated in different ways: A and B are generated independently by annotators and P is a paraphrased as-much-as-possible version of A. 
Our goal is to simulate the availability of multiple human references of varying quality levels. 
As reported in the findings of WMT20 Metrics task \cite{mathur-etal-2020-results}, in realistic scenarios the available references have very disparate quality levels, and the quality of human references is not always known. We thus calculate the performance when using each of the Human-A, Human-B and Human-P references individually, and then compare randomly sampling $r$ from $\mathcal{R}$ with averaging predictions over each $r$ in $\mathcal{R}$, hypothesizing that the combination of references will result in reduced model uncertainty. 

We can see in Table \ref{tab:multiref} \edit{that when having access to multiple references, combining all available references ($\mathrm{Mul}$)} results in narrower confidence intervals \edit{compared to sampling single references ($\mathrm{S}$-1) or even pairs of references ($\mathrm{S}$-2)} as indicated by the decreasing values in sharpness. \edit{Apart from sharpness, the model seems to benefit from the addition of new knowledge, since we see consistent improvement in performance for PPS and NLL metrics. Thus, with the incorporation of additional human references we obtain models that are more confident -- and rightly so, since they are more predictive too. Combining this information with the performance of singleton reference sets in Table \ref{tab:multiref-s}, we note that even among human references, the estimated reference quality seems to have an impact both on the predictive accuracy (PPS) and confidence (UPS, NLL, Sharpness). Both for $\mathrm{S}$-$\mathrm{N}$ and $\mathrm{Mul}$ approaches, the inclusion of Human-P in the reference set results in performance drop across all metrics. Still, the negative impact of Human-P decreases with the increase of combined references and we can conclude that when there is no information on the estimated quality of references the best approach is to combine them: for $\mathcal{R}=\{A,B,P\}$, $\mathrm{Mul}$ results in  similar performance to Human-A. }


\input{multi_ref_tables}


\subsection{Detection of critical translation mistakes}
\label{sec:ir}
\begin{figure}
        \begin{minipage}[b]{.48\textwidth}
        \subfigure[Recall@N, worst 2\%]{\includegraphics[width=\textwidth]{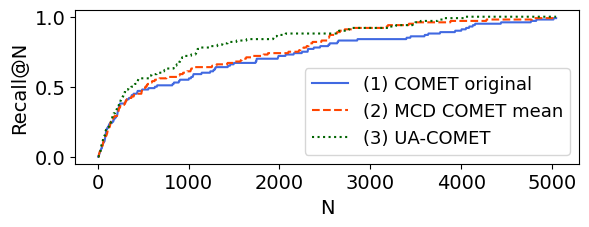}}
        \end{minipage}
        \hfill
        \begin{minipage}[b]{.48\textwidth}
        \subfigure[Precision@N, worst 2\%]{\includegraphics[width=\textwidth]{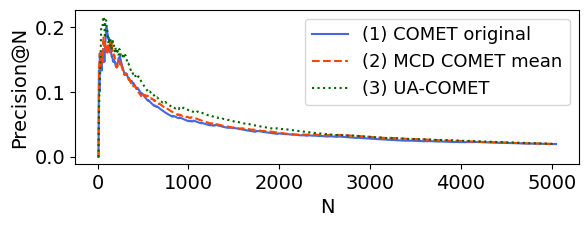}}
        \end{minipage}
        \caption{Performance on predicting the worst MTs, using \textsc{Prism} pseudo-references. The continuous (\textcolor{royalblue}{\textbf{blue}}) line corresponds to the original \textsc{Comet} prediction, while the dashed (\textcolor{internationalorange}{\textbf{orange}}) line to the averaged predictions obtained by MCD. The dotted (\textcolor{green(ncs)}{\textbf{green}}) line corresponds to predictions using the cdf \textsc{UA-Comet}.}
        \label{fig:pr-mqm}
    \end{figure}

One of the key applications where the use of uncertainty-aware MT evaluation is particularly relevant is the identification of critical translation errors that would require human assisted editing. To investigate whether uncertainty can improve performance of critical error detection, we treat the error detection as an information retrieval problem where we aim to identify the worst translations based on human annotations. We experiment with the \textsc{En-De} dataset and the corresponding MQM annotations, since MQM scores specifically designed with the distinction between major and minor translation errors in mind \cite{mqm_guidelines}. \edit{In this experiment we also take into consideration the number of words in the MT sentence and normalize scores accordingly to avoid over-penalizing for critical very long translations with accumulated minor errors. We elaborate and provide comparative examples regarding this choice in Appendix \ref{sec:app_critical}. We calculate and average the MQM scores for all 3 annotators per segment and then normalize for MT length. We then use the segments with the $n\%$ lowest scores as the retrieval targets. We present the results 
for the 2\% lowest quality segments in Figure \ref{fig:pr-mqm} and we provide additional results (with $n$ ranging from 1\% to 20\% lowest quality segments) in Appendix \ref{sec:app_critical}. We provide the statistics for the MQM data\footnote{We use a fixed dev/test split instead of k-fold cross-validation in this case. We still ensure that we do not split any document across dev/test and that  test  remains "unseen".} used in this experiment in Table \ref{tab:mqm_stats}.} Our hypothesis is that we can provide better predictions of erroneous translations, using the cumulative distribution function over $Q$ for each $\langle s, t, \mathcal{R} \rangle$ to predict the probability $P(Q \leq q_{\mathrm{err}})$, where $q_{\mathrm{err}}$ is a  quality threshold tuned on the validation set to optimize average recall@N.
We can then compare 3 ways of scoring the translations automatically: (1) using the scores $\hat{q}$ predicted by $h$ to rank translations, (2) using the mean $\hat{\mu}$ of the estimated distribution $\hat{p}_Q(q)$ instead of the single point estimate $\hat{q}$, and (3) using the uncertainty-aware parametric models to compute and rank by the probability of $q_{err}$. 

Since this scenario is more relevant to real-time/on demand translation evaluation, we test it under the assumption that there is no access to a human reference.
To handle this referenceless case ($\mathcal{R}=\varnothing$, also known as \textit{quality estimation}), we can use translations produced by an MT system outside the WMT20 participants as \textit{pseudo-references} \cite{scarton-specia-2014-document,duma-menzel-2018-benefit}. 
We use \textsc{Prism}%
\footnote{We use the m39v1 model in \url{https://github.com/thompsonb/prism} and the zero-shot translation setup.}, %
which was originally trained as a multilingual NMT model, 
\cite{thompson-post-2020-paraphrase,thompson-post-2020-automatic}.
We evaluate all scoring approaches using Recall@N and Precision@N as shown in Figure~\ref{fig:pr-mqm}. We can see that while for very small values all approaches perform similarly, the uncertainty-aware approach (\textsc{UA-Comet}) outperforms the other two for Recall as $N$ increases, while it also demonstrates higher Precision especially for small $N$ values, which are of greatest interest since we want to correct as many critical errors as possible with minimal human intervention.
\section{Conclusions}
 
We introduced \textit{uncertainty-aware} MT evaluation and showed how MT-related applications can benefit from this approach. We compared two techniques to estimate uncertainty, MC dropout and deep ensembles, across several performance indicators. 
Through experiments on three datasets with different human quality assessments encompassing several language pairs, we have shown that the resulting confidence intervals are informative and correlated with the prediction errors, leading to slightly more accurate predictions with informative uncertainty. 
Our uncertainty-aware system can take into account multiple references and becomes more confident (and accurate) when more references are available; it can so perform quality estimation without any human reference by relying on pseudo-references from other MT systems (\textsc{Prism}). We show that uncertainty-aware MT evaluation is a promising path. As a future direction, we aspire to further explore uncertainty predicting methods that tackle the different kinds of aleatoric and epistemic uncertainty described in \S\ref{sec:unc_sources} and are better tailored to the specifics of this task. 


\begin{table}[t]
\small
    \centering
    \begin{tabular}{r c c c }
    \toprule
    & \#segments & \#documents & \#MT systems\\\midrule
   dev  & 5058 & 468 & 9 \\
   test &  5049 & 468  & 9   \\
        \bottomrule
    \end{tabular}
    \caption{MQM dataset statistics for critical error detection experiments.}
    \label{tab:mqm_stats}
\end{table}

\section*{Acknowledgements}

We would like to thank Ben Peters, F\'{a}bio Kepler, Craig Stewart and the anonymous reviewers for their valuable feedback. 
This work was supported by the P2020 program MAIA (contract 045909) and Unbabel4EU (contract 042671), by the European Research Council (ERC StG DeepSPIN 758969), and by the Funda\c{c}\~{a}o para a Ci\^{e}ncia e Tecnologia through contract UIDB/50008/2020. 

\bibliography{anthology,custom}
\bibliographystyle{acl_natbib}

\clearpage
\newpage
\appendix
\input{emnlp2021_appendix}

\end{document}

%% file: table_examples.tex
\begin{table}[t]
\small
\centering
\addtolength{\tabcolsep}{-1.5pt}
\begin{tabular}{cccc}
\toprule
MT & DA & \textsc{Comet} & \textsc{UA-Comet} \\  
\midrule


\fontencoding{T2A}\selectfont{Она сказала,}
& -0.815 & \textcolor{red}{\textit{0.586}} & \textcolor{black}{0.149}   \\
\fontencoding{T2A}\selectfont{'Это не собирается}
&  &  & \textcolor{green!55!blue}{\textbf{[-0.92, 1.22]}}  \\
\fontencoding{T2A}\selectfont{работать.}
&  &  & \\[0.2cm]
\multicolumn{4}{c}{Gloss: ``\textit{She said, `that's not willing to work}''} \\
\midrule
\fontencoding{T2A}\selectfont{Она сказала:} & 0.768 & \textcolor{black}{1.047} & \textcolor{black}{1.023}  \\
\fontencoding{T2A}\selectfont{«Это не сработает.} &  &  & 
\textcolor{black}{[0.673, 1.374]}   \\[0.2cm]
\multicolumn{4}{c}{Gloss: ``\textit{She said, «That will not work}''} \\
\bottomrule
\end{tabular}
\addtolength{\tabcolsep}{1.5pt}
\caption{Example of uncertainty-aware MT evaluation for a sentence in the WMT20 dataset. Shown are two Russian translations of the same English source ``\textit{She said, `That's not going to work.}'' with reference  ``\fontencoding{T2A}\selectfont{Она сказала: ``Не получится.}'' \fontencoding{T1}\selectfont{For the first sentence, \textsc{Comet} provides a point estimate (in \textcolor{red}{\textit{red}}) that overestimates quality, as compared to a human direct assessment (DA), while our \textsc{UA-Comet} (in \textcolor{green!55!blue}{\textbf{green}}) returns a large 95\% confidence interval  which contains the DA value. For the second sentence \textsc{UA-Comet} is confident and returns a narrow 95\% confidence interval.}} 
\label{tab:examples}
\end{table}

%% file: table_WMT20.tex
\begin{table}[t]
\small
\centering
\addtolength{\tabcolsep}{-0.5pt}
\resizebox{7.7cm}{!}{
\begin{tabular}{ccccccc}
\toprule
&  & PPS $\uparrow$ & UPS $\uparrow$ & NLL $\downarrow$   & ECE $\downarrow$ & Sha. $\downarrow$ \\  
\midrule
\multirow{3}{*}{\rotatebox{90}{\textsc{En-De}}} & MCD   & 0.576 & \underline{0.284} & \underline{1.330} & \underline{0.014} & 0.645    \\
  & DE  & \underline{0.581}  & 0.246  & 1.364  &  0.023  & \underline{0.523} \\
   & Basel. & 0.576 & - & 1.337 & 0.079 & 0.845   \\  \midrule
\multirow{3}{*}{\rotatebox{90}{\textsc{En-Zh}}}  & MCD   & 0.333   & 0.064   & 1.779 & 0.024  & \underline{0.701}  \\
  & DE   & \underline{0.354}   & \underline{0.477}   & \underline{1.435} & \underline{0.020}  & 0.762  \\
  & Basel. & 0.329 & - & 1.570 & 0.090 & 1.342  \\  \midrule
\multirow{3}{*}{\rotatebox{90}{\textsc{En-Ta}}}  & MCD   & 0.658   & 0.015   & 1.226 & 0.022  & 0.585  \\
  & DE & \underline{0.675}   & \underline{0.068}   & \underline{1.200} & \underline{0.018}  & \underline{0.564}  \\
  & Basel. & 0.655 & - & 1.237 & 0.028 & 0.691  \\  \midrule
\multirow{3}{*}{\rotatebox{90}{\textsc{Zh-En}}}  & MCD   & 0.314   & 0.109   & 1.628 & \underline{0.015}  & 0.971   \\
  & DE  & \underline{0.319}   & \underline{0.174}   & 1.591 & 0.016  & \underline{0.928}   \\
  & Basel. & 0.313 & - & \underline{1.580} & 0.059 & 1.374    \\  \midrule
\multirow{3}{*}{\rotatebox{90}{\textsc{En-Ja}}}  & MCD   & 0.640   & \underline{0.165}   & 1.237 & \underline{0.011}  & 0.591   \\
  & DE  & \underline{0.651}   & 0.093   & \underline{1.225} & 0.015  &\underline{0.556} \\
  & Basel. & 0.636 & - & 1.259 & 0.035 & 0.725   \\  \midrule
\multirow{3}{*}{\rotatebox{90}{\textsc{En-Cs}}}  & MCD   & 0.691   & \underline{0.207}   & 1.163 & \underline{0.013}  & 0.548   \\
  & DE  & \underline{0.729}   & 0.163   & \underline{1.100} & \underline{0.013}  & \underline{0.455}   \\
  & Basel. & 0.695 & - & 1.172 & 0.036 & 0.608 \\  \midrule
\multirow{3}{*}{\rotatebox{90}{\textsc{En-Ru}}}  & MCD   & 0.536   & \underline{0.142}   & 1.378 & \underline{0.021}  & 0.767   \\
  & DE  & \underline{0.578}   & 0.139   & \underline{1.320} & 0.023  & \underline{0.670}   \\
  & Basel. & 0.532 & - & 1.383 & 0.041 & 0.925 \\  \midrule
\multirow{3}{*}{\rotatebox{90}{\textsc{En-Pl}}}  & MCD   & 0.611   & \underline{0.199}   & 1.275 & 0.015  & 0.650   \\
  & DE  & \underline{0.650}   & 0.176   & \underline{1.224} & \underline{0.012}  & \underline{0.581}  \\
  & Basel. & 0.608 & - & 1.301 & 0.042 & 0.783   \\  \midrule
\multirow{3}{*}{\rotatebox{90}{\textsc{En-Iu}}}  & MCD   & 0.300   & 0.223  & 1.600 & \underline{0.016}  & \underline{1.016}  \\
 & DE  & \underline{0.308}  & \underline{0.319}  & 1.682 & 0.026  & 1.052 \\
 & Basel. & 0.292 & - & \underline{1.594} & 0.077 & 1.410  \\ 
\bottomrule
\end{tabular}
}
\addtolength{\tabcolsep}{-0.5pt}
\caption{Results for segment-level DA prediction. \underline{Underlined} numbers indicate the best result for each language pair and evaluation metric. 
Reported are the predictive Pearson score $r(\hat{\mu},q^*)$ (PPS), the uncertainty Pearson score $r(|q^*-\hat{\mu}|,\hat{\sigma})$ (UPS), the negative  log-likelihood (NLL), the expected calibration error (ECE), and the sharpness (Sha.) Note that the UPS of the baseline is always zero, since it has a fixed variance.}
\label{tab:da_seg}
\end{table}


%% file: table_MQM.tex
\begin{table}[t]
\small
\centering
\addtolength{\tabcolsep}{-0.5pt}
\resizebox{7.7cm}{!}{
\begin{tabular}{ccccccc}
\toprule
&  & PPS $\uparrow$ & UPS $\uparrow$ & NLL $\downarrow$   & ECE $\downarrow$ & Sha. $\downarrow$ \\  
\midrule
\multirow{3}{*}{\rotatebox{90}{\textsc{En-De}}} 
& MCD & 0.452 & \underline{0.409} & \underline{1.433} & \underline{0.024} & 0.674\\
& DE & \underline{0.459} & 0.336 & 1.435 & 0.035  & \underline{0.556} \\
& Basel. & 0.452 & - & 1.437 & 0.094 & 1.031\\
\midrule
\multirow{3}{*}{\rotatebox{90}{\textsc{Zh-En}}}  
&MCD & \underline{0.503} & \underline{0.309} & \underline{1.402} & \underline{0.018} & 0.721\\
& DE & 0.485 & 0.257 & 1.415 & 0.023 & \underline{0.653} \\
& Basel. & 0.503 & - & 1.398 & 0.059 & 0.953\\
\bottomrule
\end{tabular}
}
\addtolength{\tabcolsep}{-0.5pt}
\caption{Results for segment-level MQM prediction. \underline{Underlined} numbers indicate the best result for each language pair and evaluation metric.}
\label{tab:mqm_seg}
\end{table}

%% file: table_QT21.tex
\begin{table}[t]
\small
	\centering
	\addtolength{\tabcolsep}{-0.5pt}
	\resizebox{7.7cm}{!}{
    \begin{tabular}{ccccccc}
    \toprule
  &  & PPS $\uparrow$ & UPS $\uparrow$ & NLL $\downarrow$   & ECE $\downarrow$ & Sha. $\downarrow$ \\  
    \midrule
    \multirow{3}{*}{\rotatebox{90}{\textsc{En-De}}} & MCD & \underline{0.765} & 0.384 & 1.054 & 0.023 & \underline{0.325}  \\
    & DE & 0.703 & \underline{0.408} & 1.110 & \underline{0.017} & 0.406  \\
    & Basel. & 0.761 & - & \underline{1.052} & 0.120 & 0.478  \\
    \midrule
    \multirow{3}{*}{\rotatebox{90}{\textsc{De-En}}}  & MCD & \underline{0.769} & 0.475 & \underline{0.964} & \underline{0.029} & \underline{0.329}  \\
    & DE & 0.702 & \underline{0.498} & 1.100 & 0.040 & 0.330  \\
    & Basel. & 0.767 & - & 1.046 & 0.140 & 0.469 \\
    \midrule
    \multirow{3}{*}{\rotatebox{90}{\textsc{En-Lv}}}  & MCD & \underline{0.778} & 0.376 & 1.209 & \underline{0.020} & \underline{0.284} \\
    & DE & 0.709 & \underline{0.377}  & 1.064 & 0.022 & 0.328 \\
    & Basel. & 0.772 & - & \underline{1.017} & 0.108 & 0.454 \\
    \midrule
    \multirow{3}{*}{\rotatebox{90}{\textsc{En-Cs}}}  & MCD & \underline{0.753} & 0.173 & 1.097 & 0.038 & \underline{0.413} \\
    & DE & 0.672 & \underline{0.216} & 1.222 & \underline{0.024} & 0.536  \\
    & Basel. & 0.752 & - & \underline{1.076} & 0.050 & 0.498  \\
    \bottomrule
    \end{tabular}
    }
	\addtolength{\tabcolsep}{0.5pt}
    \caption{Results for segment-level HTER prediction in QT21. \underline{Underlined} numbers indicate the best result for each language pair and evaluation metric.}
	\label{tab:hter_seg}
\end{table}

%% file: multi_ref_tables.tex
\begin{table}[h]
\small
    \centering
    \addtolength{\tabcolsep}{-0.9pt}
    \begin{tabular}{ c c c  c c c c}
    \toprule
          &  \#$r$  & PPS $\uparrow$  & UPS $\uparrow$ & NLL $\downarrow$ & ECE $\downarrow$ &  Sha. $\downarrow$ \\
         \midrule
         \multicolumn{7}{c}{$\mathcal{R}$=\{A,B\}}\\
         \midrule
         $\mathrm{S}$-1 & 1 & 0.452 & 0.407 & 1.403 & \underline{0.017} & 0.746  \\
	    $\mathrm{Mul}$  & 2 & \underline{0.471} & \underline{0.389} & \underline{1.388} & 0.020 & \underline{0.718}  \\
         \midrule
         \multicolumn{7}{c}{$\mathcal{R}$=\{B,P\}}\\
         \midrule
           $\mathrm{S}$-1 &1 & 0.391 & 0.327 & 1.470 & 0.029 & 0.837  \\
	    $\mathrm{Mul} $ & 2 &\underline{0.441} & \underline{0.331} & \underline{1.429} & \underline{0.013} & \underline{0.753}  \\
         \midrule
         \multicolumn{7}{c}{$\mathcal{R}$=\{A,P\}}\\
         \midrule
           $\mathrm{S}$-1 &1 & 0.406 & 0.334 & 1.475 & 0.026 & 0.852  \\ 
	    $\mathrm{Mul} $	& 2 & \underline{0.433} & \underline{0.339} & \underline{1.460} & \underline{0.019} & \underline{0.719}  \\
         \midrule
         \multicolumn{7}{c}{$\mathcal{R}$=\{A,B,P\}}\\
         \midrule
        $\mathrm{S}$-1 & 1 & 0.402 & \underline{0.355} & 1.473 & 0.026 & 0.825 \\
        $\mathrm{S}$-2 & 2 & 0.441 & 0.348 & 1.424 & 0.019 & 0.756  \\
        $\mathrm{Mul}$ & 3 & \underline{0.455} & 0.351 & \underline{1.417} & \underline{0.018} & \underline{0.702} \\
         \bottomrule
    \end{tabular}
    \addtolength{\tabcolsep}{-0.9pt}
    \caption{Performance over multiple references and combination patterns on \textsc{en-de} Google MQM annotations. $\mathrm{S}$-$\mathrm{N}$ signifies sampling w/o replacement $\mathrm{N}$ references from $\mathcal{R}$; $\mathrm{Mul}$ signifies combining estimates over multiple references in $\mathcal{R}$ as described in \S\ref{sec:refs}. \underline{Underlined} numbers indicate the best result for each evaluation metric and reference set. 
    }
    \label{tab:multiref}
\end{table}

\begin{table}[h]
\small
    \centering
    \addtolength{\tabcolsep}{-0.9pt}
    \begin{tabular}{ c  c c c c c}
    \toprule
          &    PPS $\uparrow$  & UPS $\uparrow$ & NLL $\downarrow$ & ECE $\downarrow$ &  Sha. $\downarrow$ \\
          \midrule
         $\mathcal{R}$=\{A\}& \underline{0.452} & \underline{0.409} & \underline{1.433} & 0.024 & \underline{0.674} \\
         $\mathcal{R}$=\{B\}&  0.442 & 0.400 & 1.406 &  \underline{0.015} & 0.782  \\
         $\mathcal{R}$=\{P\}& 0.391 & 0.275 & 1.511 & 0.020 & 0.783  \\
         \bottomrule
    \end{tabular}
    \addtolength{\tabcolsep}{-0.9pt}
    \caption{Performance over singleton reference sets on \textsc{en-de} Google MQM annotations. \underline{Underlined} numbers indicate the best result for each evaluation metric. 
    }
    \label{tab:multiref-s}
\end{table}

%% file: emnlp2021_appendix.tex
\section{Baseline with Fixed Variance}
\label{sec:optimal_fixed_variance}

We show here that, when $\hat{p}_Q(q) = \mathcal{N}(q, \hat{\mu}, \hat{\sigma}^2)$ is a Gaussian distribution, the optimal fixed variance that minimizes NLL is 
\begin{equation*}
    \sigma_{\mathrm{fixed}}^2 = \frac{1}{|\mathcal{D}|} \sum_{j = 1}^{|\mathcal{D}|} (q_j^* - \hat{\mu}_j)^2.
\end{equation*}
To show this, observe that 
\begin{align*}
&\sigma_{\mathrm{fixed}}^2 = \arg\min_{\sigma^2} -\sum_{j = 1}^{|\mathcal{D}|} \log \mathcal{N}(q_j^*, \hat{\mu}_j, \sigma^2)\nonumber\\
&= \arg\min_{\sigma^2} \sum_{j = 1}^{|\mathcal{D}|} \left(  \frac{\log (2\pi\sigma^2)}{2} + \frac{(q^*_j + \hat{\mu}_j)^2}{2\sigma^2} \right)\nonumber\\
&= \arg\min_{y > 0} \underbrace{ \sum_{j = 1}^{|\mathcal{D}|} \left( -\frac{\log (\pi^{-1} y)}{2} + (q^*_j + \hat{\mu}_j)^2 y \right)}_{:= F(y)},
\end{align*}
where we made the variable substitution $y = \frac{1}{2\sigma^2}$ and we defined the function $F:\mathbb{R}_{>0} \rightarrow \mathbb{R}$, which is convex on its domain and tends to $+\infty$ when $y\rightarrow 0_+$ and when $y \rightarrow +\infty$, hence it has a global minimum. 
Equating the derivative of the objective function to zero, we get
\begin{align*}
& 0 = F'(y) 
= -\frac{|\mathcal{D}|}{2y} + \sum_{j=1}^{|\mathcal{D}|}(q^*_j - \hat{\mu}_j)^2,
\end{align*}
from which we get 
\begin{align*}
y = \left(\frac{2}{|\mathcal{D}|}\sum_{j=1}^{|\mathcal{D}|}(q^*_j - \hat{\mu}_j)^2\right)^{-1}
\end{align*}
and 
$\sigma^2 = \frac{1}{2y} = \frac{1}{|\mathcal{D}|}\sum_{j=1}^{|\mathcal{D}|}(q^*_j - \hat{\mu}_j)^2$ as desired.

\section{Datasets}
\label{sec:datasets}

We present in Table \ref{tab:table_newstest20} descriptive statistics of datasets used in our experiments.

\begin{table}[!ht]
	\centering
	\small
	\begin{tabular}{*{3}{c} c  }
		\toprule
		  &  WMT20 & QT21 & Google \\
		\midrule
		avg \# seg per LP & 1391 & 1000  & 1709   \\  
		avg \# doc  & 74 & -  & 99 \\ 
		max \# systems per LP & 16 & 2  & 8  \\
		avg doc length & 16 & - & 12 \\
		\# LPs & 9  & 4  & 2  \\
		annotations & DA & HTER & MQM \\
		 \bottomrule
	\end{tabular}
	\caption{Descriptive statistics of the newstest2020 datasets. Systems Human-A, Human-B and Human-P are excluded. Google corresponds to the MQM extension on the WMT20 dataset.
	}
	\label{tab:table_newstest20}
\end{table}

\begin{figure}[!ht]
\includegraphics[width=\columnwidth]{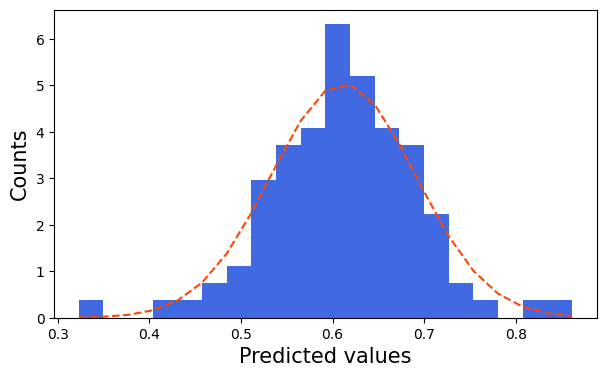}
\caption{Distribution of predicted values for a random sample from WMT20 dataset, \textsc{En-Ta} language pair.}
\label{fig:gaussian-data}
\end{figure}

In Fig.\ref{fig:gaussian-data} we show the distribution of predicted quality estimates for a random sample from WMT20 dataset, (\textsc{En-Ta} language pair\footnote{Based on a translation produced by the OPPO system, for the segment with index 473 (randomly sampled).}), with the corresponding superimposed gaussian to demonstrate the perceived fit.

\section{Hyperparameter Tuning}
\label{sec:hyperparams}
The number of dropout runs was tuned on the $[25,200]$ interval with a step of 25 on the \textsc{En-De} WMT20 data. We show the results in Table~\ref{tab:dropout}.
In preliminary experiments, we found that increasing the dropout probability beyond $0.1$ did not bring any gains, therefore we used this number.  We also found that dropping only the feed-forward layers of \textsc{Comet} and/or the pooling layers was ineffective, therefore we applied dropout on all \textsc{Comet} layers for all experiments presented in this paper. 

\input{table_MCDruns}


\section{Deep Ensembles}

\begin{table*}[ht!]
\small
\centering
\begin{tabular}{lcc}
\toprule 
\textbf{Hyperparameter}  & \textbf{HTER}  & \textbf{DA}\tabularnewline
\midrule 
Encoder Model  & XLM-R (base)  & XLM-R (large) \tabularnewline
Optimizer  & Adam  & Adam \tabularnewline
nº frozen epochs  & 1  & 0.4 \tabularnewline
Learning rate  & 3e-05  & 3e-04 \tabularnewline
Encoder Learning Rate  & 1e-05  & 1e-05 \tabularnewline
Layerwise Decay  & 0.95  & 0.95 \tabularnewline
Batch size  & 16  & 16 \tabularnewline
Loss function  & Mean squared error  & Mean squared error \tabularnewline
Dropout  & 0.1  & 0.1 \tabularnewline
Hidden sizes  & [3072, 1536]  & [3072, 1536] \tabularnewline
Encoder Embedding layer & Frozen & Frozen \tabularnewline
FP precision  & 32  & 32 \tabularnewline
Nº Epochs & 2 & 2 \\
\bottomrule
\end{tabular}
\caption{Hyperparameters used to train the deep ensembles.} 
\label{tab:hp_ensembles}
\end{table*}

Table \ref{tab:hp_ensembles} shows the hyperparameters used to train the DA and HTER estimators for our deep ensembles. In both cases we trained 4 models with 
different seeds 
and used as fifth model the \textit{wmt-large-da-estimator-1719} and the \textit{wmt-large-hter-estimator} available in \url{https://github.com/Unbabel/COMET}. Each of these models has 583M parameters and were trained on a single Nvidia Quadro RTX 8000 GPU\footnote{\url{https://www.nvidia.com/en-us/design-visualization/quadro/rtx-8000/}} for $\approx34$ and $\approx3.5$ hours for the DA models and HTER models, respectively. 
Regarding the validation performance recorded during training, the DA models achieve a PPS of $0.612\pm0.002$, while the HTER models achieve a PPS of $0.663\pm0.012$. 

\section{Non-parametric Estimation of Confidence Intervals}
\label{sec:non-parametric}


The parametric Gaussian approach we chose to 
obtain confidence intervals, described in \S\ref{sec:uncertainty}, fits relatively well our data (see Figure~\ref{fig:gaussian-data}). 
However, this approach  makes a strong assumption about the shape of $\hat{p}_Q(q)$, and therefore we experimented also with a non-parametric bootstrapping technique to estimate confidence intervals. Such approach has been successful in several NLP tasks \citep{koehn-2004-statistical, li2017modeling, ye-etal-2021-towards}. 
In this case, we construct the confidence intervals $I(\gamma)$ by using the percentile method \cite{efron1992bootstrap, johnson2001introduction}.  
We take the range of point estimates in $\mathcal{Q}$ that cover equal $\frac{\gamma}{2}$ proportions around the median of the $\hat{p}_Q(q)$ distribution as the desired confidence interval, represented by the corresponding sample quantiles. 
Since this approach typically require many samples to obtain accurate estimates of the quantiles, we left out the deep ensemble method from this experiment (which would require training too many models) and focused only on samples obtained from MC dropout, using $M=100$ as in the parametric Gaussian experiments. 

Since this approach does not produce a full distribution $\hat{p}_Q(q)$ but only the median $\hat{\mu}_{\mathrm{med}}$ and confidence intervals $I(\gamma)$, the evaluation metrics UPS, NLL, and sharpness cannot be directly applied. Therefore, we evaluated with the following modifications of predictive Pearson score and ECE.

\paragraph{Predictive Pearson score}
For Pearson-related evaluation we use the PPS performance indicator defined in \S~\ref{sec:pearson_correlations}, but we measure the correlation between groundtruth quality scores $q^*$ and the median $\hat{\mu}_{\mathrm{med}}$, instead of the average $\hat{\mu}$.

\paragraph{Calibration Error} 

To compute ECE we use the same method as defined in Eq.~\ref{eq:ece}. We use this metric with $M=20$ to assess the ability of the non-parametric method to estimate confidence intervals.

\paragraph{Experiments}

The results are shown in Table~\ref{tab:da_seg_np}. 
Overall, MC dropout outperforms the baseline across both measures (except for PPS in \textsc{En-Cs}) but the improvement is marginal. The performance of the parametric approach for the same dataset in Table \ref{tab:da_seg} is better than non-parametric for both reported ECE and PPS. Still, ECE values are close to the ones obtained with the parametric approach for all language pairs, and we can obtain a well-calibrated model with the non-parametric approach too (compared to the baseline). 


The observed performance of a non-parametric approach could be limited by the number of observed samples and the method used to generate those (MC dropout). In \citet{ye-etal-2021-towards} a similar experiment of confidence intervals calibration was performed over $1000$ bootstrapped samples. Running this number of MC dropout runs would be very expensive in practice and out of scope of this work.

\begin{table}[t]
\small
\centering
\begin{tabular}{ cccc }\toprule
  &  & PPS $\uparrow$  &  ECE $\downarrow$  \\  \midrule
\multirow{3}{*}{\textsc{En-De}}
   & MC dropout & 0.576 & \underline{0.016} \\
   & Baseline & 0.576 & 0.071   \\  \midrule
\multirow{3}{*}{\textsc{En-Zh}}  & MC dropout & \underline{0.332} & \underline{0.030}  \\
  & Baseline & 0.329 & 0.062  \\  \midrule
\multirow{3}{*}{\textsc{En-Ta}}  & MC dropout & \underline{0.657} & \underline{0.024}  \\
  & Baseline & 0.655 & 0.050   \\  \midrule
\multirow{3}{*}{\textsc{Zh-En}}  & MC dropout & \underline{0.314} & \underline{0.016}  \\
  & Baseline & 0.313 & 0.057  \\  \midrule
\multirow{3}{*}{\textsc{En-Ja}}  & MC dropout  & \underline{0.640} & \underline{0.015}   \\
  & Baseline & 0.636 & 0.051 \\  \midrule
\multirow{3}{*}{\textsc{En-Cs}}  & MC dropout  & 0.691 & \underline{0.013}  \\
  & Baseline & \underline{0.695} & 0.053  \\  \midrule
\multirow{3}{*}{\textsc{En-Ru}}  & MC dropout & \underline{0.536} & \underline{0.019} \\
  & Baseline  & 0.532  & 0.061   \\  \midrule
\multirow{3}{*}{\textsc{En-Pl}}  & MC dropout & \underline{0.611} & \underline{0.016} \\
  & Baseline & 0.608 & 0.052  \\  \midrule
\multirow{3}{*}{\textsc{En-Iu}}  & MC dropout & \underline{0.300} & \underline{0.016} \\
 & Baseline & 0.292 & 0.057 \\ \bottomrule
 \end{tabular}
\caption{Results for segment-level DA prediction for a non-parametric approach. \underline{Underlined} numbers indicate the best result for each language pair and evaluation metric. 
Reported are the predictive Pearson score $r(\hat{\mu}_{\mathrm{med}}, q^*)$ (PPS), where $\hat{\mu}_{\mathrm{med}}$ is the median, and the expected calibration error (ECE).}
\label{tab:da_seg_np}
\end{table}

\section{Detection of Critical Translation Mistakes}
\label{sec:app_critical}
\begin{figure*}
        \begin{minipage}[b]{.48\textwidth}
        \subfigure[Recall@N, worst 1\%]{\label{fig:R-a}\includegraphics[width=\textwidth]{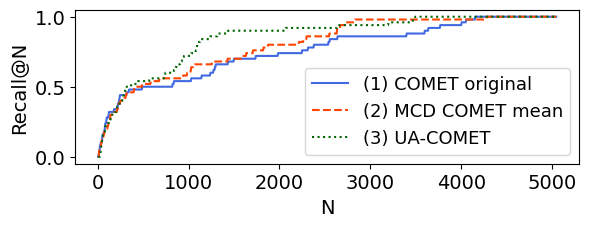}}
        \subfigure[Recall@N, worst 2\%]{\label{fig:R-b}\includegraphics[width=\textwidth]{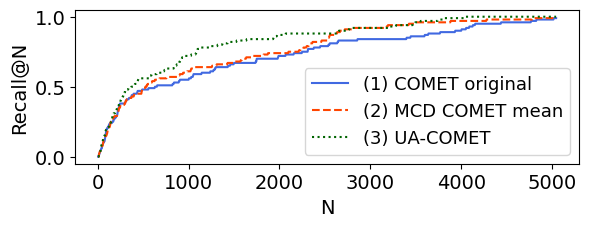}}
        \subfigure[Recall@N, worst 5\%]{\label{fig:R-c}\includegraphics[width=\textwidth]{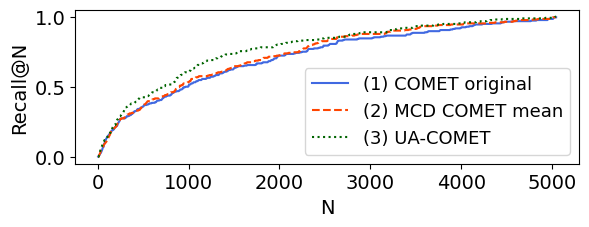}}
        \subfigure[Recall@N, worst 10\%]{\label{fig:R-d}\includegraphics[width=\textwidth]{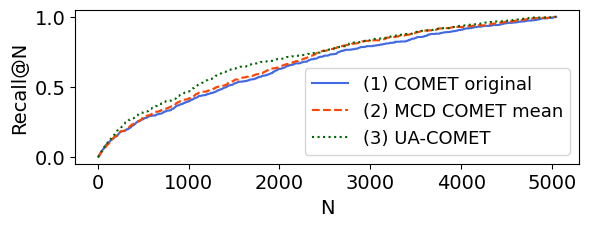}}
        \subfigure[Recall@N, worst 15\%]{\label{fig:R-e}\includegraphics[width=\textwidth]{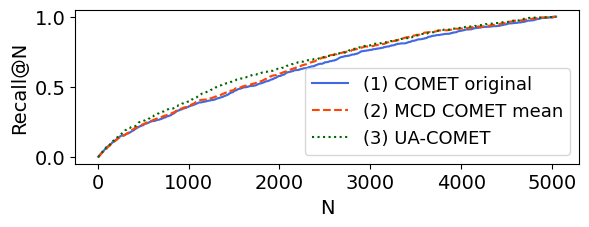}}
        \subfigure[Recall@N, worst 20\%]{\label{fig:R-f}\includegraphics[width=\textwidth]{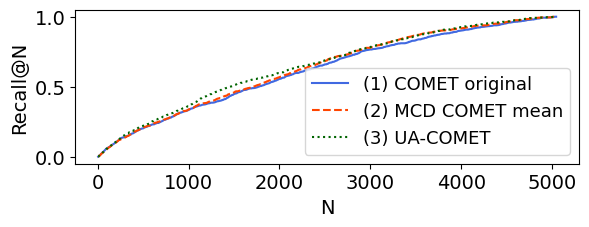}}
        \end{minipage}
        \hfill
        \begin{minipage}[b]{.48\textwidth}
        \subfigure[Precision@N, worst 1\%]{\label{fig:P-a}\includegraphics[width=\textwidth]{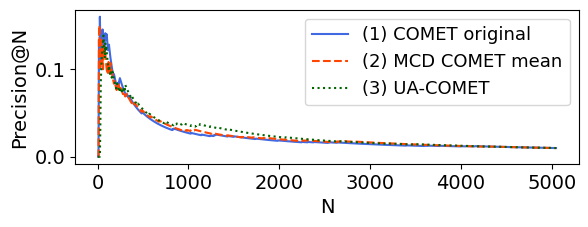}}
        \subfigure[Precision@N, worst 2\%]{\label{fig:P-b}\includegraphics[width=\textwidth]{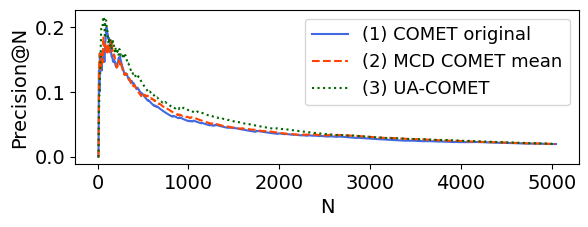}}
        \subfigure[Precision@N, worst 5\%]{\label{fig:P-c}\includegraphics[width=\textwidth]{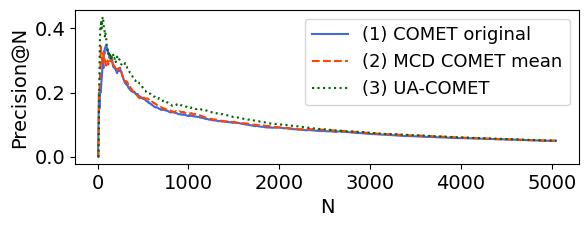}}
        \subfigure[Precision@N, worst 10\%]{\label{fig:P-d}\includegraphics[width=\textwidth]{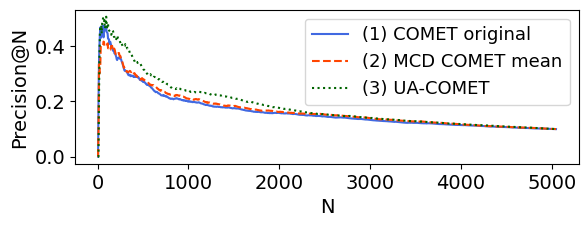}}
        \subfigure[Precision@N, worst 15\%]{\label{fig:P-e}\includegraphics[width=\textwidth]{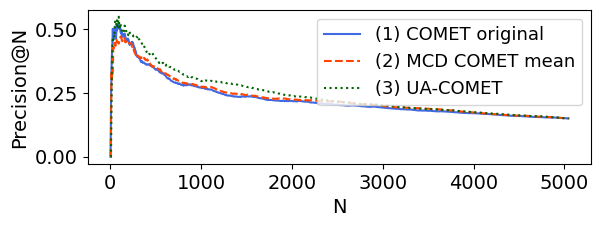}}
        \subfigure[Precision@N, worst 20\%]{\label{fig:P-f}\includegraphics[width=\textwidth]{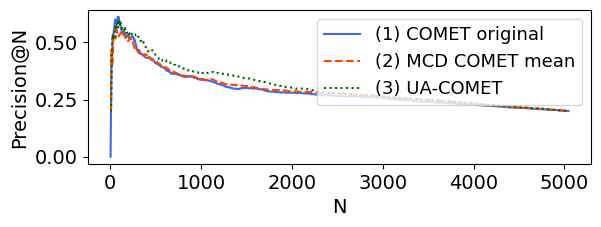}}
        \end{minipage}
        \caption{Performance on predicting the worst MTs, using \textsc{Prism} pseudo-references. The continuous (\textcolor{royalblue}{\textbf{blue}}) line corresponds to the original \textsc{Comet} prediction, while the dashed (\textcolor{internationalorange}{\textbf{orange}}) line to the averaged predictions obtained by MCD. The dotted (\textcolor{green(ncs)}{\textbf{green}}) line corresponds to predictions using the cdf \textsc{UA-Comet}.}
        \label{fig:pr-mqm-ext}
    \end{figure*}

\edit{We provide more detailed experiments of the critical translation error detection in Figure \ref{fig:pr-mqm-ext}, showing the Recall@N and Precision@N for different error proportions from  the dataset, ranging from $1\%$ to $20\%$. We can see that while increasing the proportion of errors considered critical, the Recall@N performance gap for \textsc{UA-Comet} and \textsc{Comet} decreases. }

We show examples of the worst translations according to MQM scores with and without length normalisation in Tables \ref{tab:mqm_bottom_20_len} and \ref{tab:mqm_bottom_20} respectively, in order to better demonstrate the impact of length normalisation on the selection of critical errors.
\input{tables_mqm_sort}





%% file: table_MCDruns.tex
\begin{table}[!h]
\small
\addtolength{\tabcolsep}{-0.5pt}
\begin{tabular}{cccccc}
\toprule
  \# runs  & PPS $\uparrow$ & UPS $\uparrow$ & NLL $\downarrow$ & ECE $\downarrow$ & Sharp. $\downarrow$ \\ 
\midrule
 25 & 0.580 & 0.200 & 1.346 & 0.015 & 0.657 \\
 50 & 0.581 & 0.204 & 1.334 & 0.015 & 0.635 \\
 75 & 0.581 & 0.204 & 1.328 & 0.014 & 0.627  \\
 \rowcolor{palecornflowerblue}
100 & 0.582 & 0.206 & 1.323 & 0.014 & 0.624\\
 125 & 0.582 & 0.207 & 1.326 & 0.014 & 0.636 \\
 150 & 0.582 & 0.209 & 1.323 & 0.014 & 0.631 \\
 175 & 0.582 & 0.209 & 1.324 & 0.014 & 0.633 \\
 200 & 0.582 & 0.210 & 1.322 & 0.015 & 0.623 \\
\bottomrule
\end{tabular}
\addtolength{\tabcolsep}{-0.5pt}
\caption{[DA] Segment-level results obtained with different number of dropout runs.}
\label{tab:dropout}
\end{table}

%% file: tables_mqm_sort.tex
\begin{table*}
\small
\resizebox{\textwidth}{!}{%
\begin{tabular}{@{}p{9cm}p{9cm}l@{}}
\toprule
\textbf{source sentences} &
  \textbf{MT sentences} &
  \textbf{MQM} \\ \midrule
  \rowcolor{LightCyan}
Vulnerable Dems air impeachment concerns to Pelosi &
  Anfällige Dems Luft Amtsenthebungsbedenken an Pelosi &
  17.67 \\
  \rowcolor{LightCyan}
Vulnerable Dems air impeachment concerns to Pelosi &
  Anfällige Dems Luft-Impeachment Bedenken gegen Pelosi &
  17.33 \\
  \rowcolor{LightCyan}
Vulnerable Dems air impeachment concerns to Pelosi &
  Verletzliche Dems-Luft-Impeachment-Bedenken gegen Pelosi &
  17.67 \\
Government Retires 15 More Senior Tax Officials On Graft Charges &
  Regierung scheidet aus 15 weiteren hohen Steuerbeamten wegen Graft-Gebühren aus &
  17 \\
  \rowcolor{LightCyan}
Hideous' Central Coast camouflage child rapist ordered to look at victim in court &
  "Hideous" Central Coast Tarnung Kindervergewaltiger bestellt, um Opfer vor Gericht zu betrachten &
  20.07 \\
A third wrote: "Don't fall for it Khloe." &
  Ein dritter schrieb: „Fallen Sie nicht für Khloe.“ &
  10.37 \\
  \rowcolor{LightCyan}
Parents of 5-month-old stuffed in suitcase and tossed in dumpster get 6 years in prison &
  Eltern von 5 Monaten in Koffer gestopft und in Müllcontainer geworfen bekommen 6 Jahre Gefängnis &
  18.67 \\
The Who STOP concert last night: Friday and Sunday shows CANCELLED &
  Das Who STOP Konzert gestern Abend: Freitag und Sonntag zeigt CANCELLED &
  13.67 \\
  \rowcolor{LightCyan}
Parents of 5-month-old stuffed in suitcase and tossed in dumpster get 6 years in prison &
  Eltern von 5-Monats-Alt in Koffer gefüllt und in Mülleimer geworfen bekommen 6 Jahre im Gefängnis &
  18.67 \\
Vulnerable Dems air impeachment concerns to Pelosi &
  Vulnerable Dems Air Impeachment Bedenken für Pelosi &
  9.67 \\
Brother Jailed For Life For Pakistan Social Media Star Qandeel Baloch's Honour Killing &
  Bruder für Leben für Pakistan Social Media Star Qandeel Baloch s Ehre Tötung inhaftiert &
  15.37 \\
Vulnerable Dems air impeachment concerns to Pelosi &
  Vulnerable Dems Air Impeachment Bedenken gegen Pelosi &
  9.33\\
"I can't help the way I'm made," Whitehurst told the Sun. &
  „Ich kann nicht anders, wie ich gemacht bin“, sagte Whitehurst der Sonne. &
  12.67 \\
"I can't help the way I'm made," Whitehurst told the Sun. &
  "Ich kann nicht anders, als ich gemacht bin", sagte Whitehurst der Sonne. &
  12.4 \\
  \rowcolor{LightCyan}
Parents of 5-month-old stuffed in suitcase and tossed in dumpster get 6 years in prison &
  Eltern von 5 Monaten, die in Koffer gestopft und in Müllcontainer geworfen werden, bekommen 6 Jahre Gefängnis &
  18.33 \\
Woman STRIPS TO NOTHING in Walmart to prove she didn't steal &
  Frau STRIPS TO NOTHING in Walmart zu beweisen, dass sie nicht stehlen &
  11.33\\
Brother Jailed For Life For Pakistan Social Media Star Qandeel Baloch's Honour Killing &
  Bruder lebenslang für Pakistan eingesperrt Social Media Star Qandeel Balochs Ehrenmord &
  14.03 \\
  \rowcolor{LightCyan}
Sacramento police also announced Thursday their internal investigation did not find any policy or training violations. &
  Sacramento Polizei kündigte auch am Donnerstag ihre internen Ermittlungen fand keine Richtlinien oder Trainingsverstöße. &
  18 \\
Man pleads guilty in kidnap, torture plot of plastic surgeon &
  Mann bekennt sich schuldig bei Entführung, Folter des plastischen Chirurgen &
  11 \\
  \bottomrule
\end{tabular}%
}
\caption{Worst 20 translations according to MQM scores (averaged over 3 annotators) for \textsc{En-De}, normalised by sentence length (word number). Highlighted rows are common in both ranking approaches.}
\label{tab:mqm_bottom_20_len}
\end{table*}

\begin{table*}
\small
\resizebox{\textwidth}{!}{%
\begin{tabular}{@{}p{9cm}p{9cm}c@{}}
\toprule
\textbf{source sentences} &
  \textbf{MT sentences} &
  \textbf{MQM} \\ \midrule
"{[}Barr has{]} gone rogue," Pelosi told MSNBC Friday. "I think where they're going is a cover-up of a cover-up. I think it's sad, to have a Justice Department go so rogue. Well, they have been for a while. And now it just makes matters worse." &
  „{[}Barr hat{]} gegangen Schurken“, Pelosi sagte MSNBC Freitag. „Ich denke, wohin sie gehen, ist eine Vertuschung einer Vertuschung. Ich denke, es ist traurig, ein Justizministerium gehen so Schurken. Nun, sie haben für eine Weile. Und jetzt macht es die Sache nur noch schlimmer“. &
  22.33 \\
Add Lancaster of Pikeville told North Carolina Education Lottery officials he used five sets of his own numbers to buy a Cash 5 ticket with five plays for Monday night's drawing when he stopped at Wissam \& Brothers Inc. in Pikeville. &
  Fügen Sie Lancaster von Pikeville sagte North Carolina Education Lottery Beamten er fünf Sätze seiner eigenen Zahlen verwendet, um ein Cash 5 Ticket mit fünf Spielen für Montag Abend Zeichnung zu kaufen, als er bei Wissam \& Brothers Inc. in Pikeville hielt. &
  22.33 \\
  \rowcolor{LightCyan}
Hideous' Central Coast camouflage child rapist ordered to look at victim in court &
  "Hideous" Central Coast Tarnung Kindervergewaltiger bestellt, um Opfer vor Gericht zu betrachten &
  20.06 \\
Trump is singing from a similar songbook. His administration's Muslim-majority travel ban echoes the Islamophobia that often informs Modi's policymaking. Its callousness toward refugees mirrors the Indian government's disdain for the Rohingya population's suffering, and its detention camps parallel the ones the Modi regime is setting up. Trump's stirring of racial animosity is analogous to troublesome rhetoric from a number of Modi's cabinet members. &
  Trump singt aus einem ähnlichen Liederbuch. Das Reiseverbot seiner Regierung mit muslimischer Mehrheit spiegelt die Islamophobie wider, die oft Modis Politik informiert. Seine Anrufung gegenüber Flüchtlingen spiegelt die Verachtung der indischen Regierung für das Leiden der Rohingya-Bevölkerung und ihre Gefangenenlager parallel zu denen wider, die das Modi-Regime einrichtet. Trumps Aufregung rassischer Feindseligkeit ist analog zur lästigen Rhetorik einer Reihe von Modis Kabinettsmitgliedern. &
  19.67 \\
"Currently we are targeting young people 18 to 24 years. For the young people that's the age bracket we are looking at but of course any one above 18 and it's because we do not have evidence of children by the Constitution but as more evidence unfolds we are going to get there. For the men, we give the kit to the mother and they take it to the partner, key and priority populations such sex workers," Mr Geoffrey Tasi, the technical officer-in-charge of HIV testing services, said yesterday. &
  "Derzeit richten wir uns an Jugendliche im Alter von 18 bis 24 Jahren. Für die jungen Leute, die die Altersgruppe sind, die wir betrachten, aber natürlich jede über 18 und es ist, weil wir keine Beweise für Kinder durch die Verfassung haben, aber als mehr Beweise sich entfalten, werden wir dorthin gelangen. Für die Männer geben wir das Kit an die Mutter und sie bringen es dem Partner, Schlüssel- und Priorat solcher Sexarbeiterinnen", sagte Geoffrey Tasi, der für HIV-Testdienste zuständige technische Beamte, gestern. &
  19.07 \\
  \rowcolor{LightCyan}
Parents of 5-month-old stuffed in suitcase and tossed in dumpster get 6 years in prison &
  Eltern von 5-Monats-Alt in Koffer gefüllt und in Mülleimer geworfen bekommen 6 Jahre im Gefängnis &
  18.67 \\
  \rowcolor{LightCyan}
Parents of 5-month-old stuffed in suitcase and tossed in dumpster get 6 years in prison &
  Eltern von 5 Monaten in Koffer gestopft und in Müllcontainer geworfen bekommen 6 Jahre Gefängnis &
  18.67 \\
  \rowcolor{LightCyan}
Parents of 5-month-old stuffed in suitcase and tossed in dumpster get 6 years in prison &
  Eltern von 5 Monaten, die in Koffer gestopft und in Müllcontainer geworfen werden, bekommen 6 Jahre Gefängnis &
  18.33 \\
  \rowcolor{LightCyan}
Sacramento police also announced Thursday their internal investigation did not find any policy or training violations. &
  Sacramento Polizei kündigte auch am Donnerstag ihre internen Ermittlungen fand keine Richtlinien oder Trainingsverstöße. &
  18 \\
  \rowcolor{LightCyan}
Vulnerable Dems air impeachment concerns to Pelosi &
  Verletzliche Dems-Luft-Impeachment-Bedenken gegen Pelosi &
  17.67 \\
The 35-year-old star dumped the NBA player for good earlier this year after he was accused of cheating on her with family friend Jordyn Woods - having previously cheated when she was nine months pregnant with their daughter, True. &
  Der 35-jährige Star warf die NBA-Spielerin Anfang des Jahres endgültig ab, nachdem er beschuldigt wurde, sie mit Familienfreund Jordyn Woods betrogen zu haben - nachdem sie zuvor betrogen hatte, als sie im neunten Monat mit ihrer Tochter True schwanger war. &
  17.67 \\
  \rowcolor{LightCyan}
Vulnerable Dems air impeachment concerns to Pelosi &
  Anfällige Dems Luft Amtsenthebungsbedenken an Pelosi &
  17.67 \\
It comes just days after Tristan wrote: "Perfection" alongside the heart eye emojis underneath one of the reality stars other photos, which saw her modelling for Guess Jeans. &
  Es kommt nur wenige Tage, nachdem Tristan geschrieben hat: "Perfection" neben den Herzaugen-Emojis unter einem der Reality-Stars andere Fotos, die sie für Guess Jeans modellieren sah. &
  17.43 \\
"You're going out a youngster, but you've got to come back a star!" Blanks wrote in an Instagram caption on Wednesday, quoting the film "42nd Street." &
  "Du gehst als Jugendlicher aus, aber du musst einen Stern zurückkommen!" Blanks schrieb am Mittwoch in einem Instagram-Titel den Film "42nd Street". &
  17.43 \\
"Sounding more and more like the so-called whistle-blower isn't a whistle-blower at all," he tweeted. "In addition, all second-hand information that proved to be so inaccurate that there may not have been somebody else, a leaker or spy, feeding it to him or her? A partisan operative?" &
  "Immer mehr nach dem sogenannten Whistleblower zu klingen, ist überhaupt kein Whistleblower", twitterte er. "Außerdem alle Informationen aus zweiter Hand, die sich als so ungenau erwiesen haben, dass möglicherweise nicht jemand anderes, ein Leckerbissen oder ein Spion, sie ihm oder ihr gefüttert hat? Ein Partisanen-Agent?" &
  17.43 \\
"Currently, 86 per cent people living with HIV know their status; that means it leave us with 14 per cent of those living with HIV and do not know their status. So how do we now utilise that additional innovation. Really for me this is it ... how do we now move from this kit and create demand, especially for that 14 per cent that are sick and they need care and they are not getting care," Dr Atwine said. &
  "Derzeit kennen 86 Prozent der HIV-Infizierten ihren Status; Das bedeutet, dass wir bei 14 Prozent der HIV-Infizierten leben und ihren Status nicht kennen. Wie können wir nun diese zusätzliche Innovation nutzen? Wirklich für mich ist es ... Wie können wir jetzt von diesem Kit wegkommen und Nachfrage schaffen, vor allem für die 14 Prozent, die krank sind und Pflege brauchen und sie nicht versorgt werden", sagte Dr. Atwine. &
  17.4 \\
Sacramento police also announced Thursday their internal investigation did not find any policy or training violations. &
  Sacramento Polizei kündigte auch am Donnerstag ihre interne Untersuchung keine Politik oder Ausbildung Verstöße gefunden. &
  17.33 \\
"Currently we are targeting young people 18 to 24 years. For the young people that's the age bracket we are looking at but of course any one above 18 and it's because we do not have evidence of children by the Constitution but as more evidence unfolds we are going to get there. For the men, we give the kit to the mother and they take it to the partner, key and priority populations such sex workers," Mr Geoffrey Tasi, the technical officer-in-charge of HIV testing services, said yesterday. &
  „Gegenwärtig richten wir uns an junge Menschen zwischen 18 und 24 Jahren. Für die jungen Menschen ist das die Altersgruppe, die wir betrachten, aber natürlich jede über 18, und das liegt daran, dass wir keine Beweise für Kinder durch die Verfassung haben, aber wenn sich mehr Beweise entwickeln, werden wir dorthin gelangen. Für die Männer geben wir das Kit an die Mutter und sie bringen es an den Partner, Schlüssel- und Prioritätspopulationen wie Sexarbeiter“, sagte gestern Geoffrey Tasi, der zuständige technische Offizier für HIV-Tests. &
  17.33 \\
  \rowcolor{LightCyan}
Vulnerable Dems air impeachment concerns to Pelosi &
  Anfällige Dems Luft-Impeachment Bedenken gegen Pelosi &
  17.33 \\
  \bottomrule
\end{tabular}%
}
\caption{Worst 20 translations according to MQM scores (averaged over 3 annotators) for \textsc{En-De}. Highlighted rows are common in both ranking approaches.}
\label{tab:mqm_bottom_20}
\end{table*}